\documentclass[letterpaper, 10 pt, conference]{ieeeconf}  

\IEEEoverridecommandlockouts

\usepackage{graphics} 
\usepackage{epsfig} 
\usepackage{mathptmx} 
\usepackage{times} 
\usepackage{amsmath} 
\usepackage{amssymb}  
\usepackage{booktabs} 
\usepackage[table]{xcolor} 

\usepackage[hidelinks]{hyperref}
\usepackage[nameinlink]{cleveref}
\Crefname{figure}{Fig.}{Figs.}

\usepackage{graphicx} 
\usepackage{subcaption}
\usepackage{caption}
\usepackage{tikz}
\title{\LARGE \bf
A Proprioceptive-Only Benchmark for Quadruped State Estimation: ATE, RPE, and Runtime Trade-offs Between Filters and Smoothers}

\author{Ylenia Nisticò$^{1}$, Jo\~ao Carlos Virgolino Soares$^{1}$, Joan Solà$^{2}$, and Claudio Semini$^{1}$
\thanks{$^{1}$Dynamic Legged Systems (DLS) Lab, Istituto Italiano di Tecnologia, 16163 Genova, Italy
        {\tt\small ylenia.nistico@iit.it, joao.virgolino@iit.it, claudio.semini@iit.it}}
\thanks{$^{2}$Institut de Robòtica i Informàtica Industrial - CSIC, 08028 Barcelona, Spain
        {\tt\small jsola@iri.upc.edu}}
}

\begin{document}

\maketitle
\thispagestyle{empty}
\pagestyle{empty}

\begin{abstract}
We compare three state-of-the-art proprioceptive state estimators for quadruped robots: MUSE~\cite{nistico2025muse}, the Invariant Extended Kalman Filter (IEKF)~\cite{hartley2020contact}, and the Invariant Smoother (IS)~\cite{yoon2024invariant}, on the CYN-1 sequence of the GrandTour Dataset~\cite{frey2026grandtour}. Our goal is to give practitioners clear guidance on accuracy and computation time: we report long-term accuracy~(Absolute Trajectory Error, ATE), short-term accuracy (translational and rotational Relative Pose Error, RPE), and per-update computation time on a fixed hardware/software stack. On this dataset, RPEs are broadly similar across methods, while IEKF and IS achieve a lower ATE than MUSE. 
Runtime results highlight the accuracy-latency trade-offs across the three approaches. In the discussion, we outline the evaluation choices used to ensure a fair comparison and analyze factors that influence short-horizon metrics. Overall, this study provides a concise snapshot of accuracy and cost, helping readers choose an estimator that fits their application constraints, with all evaluation code and documentation released open-source at \url{https://github.com/iit-DLSLab/state_estimation_benchmark} for full reproducibility.
\looseness=-1
\end{abstract}

\section{INTRODUCTION}
Reliable proprioceptive state estimation is essential for agile quadruped locomotion. When exteroceptive sensing is unreliable or unavailable (e.g., due to dust/fog, long featureless corridors, or GPS outages), robots must rely on IMU, joint encoders, and contact information to estimate body motion. The challenge is twofold: (\textit{i}) maintain \emph{long-term consistency} under drift and modeling errors, and (\textit{ii}) preserve \emph{short-horizon fidelity} that directly affects feedback control and foothold planning, while meeting tight latency and compute budgets on embedded platforms.

Early work, such as~\cite{bloesch2013state}, established that IMU and leg kinematics can yield reliable odometry when fused consistently. The Two-State Implicit Filter (TSIF)~\cite{bloesch2017two} further simplified the estimator structure to improve robustness and real-time performance. Modern modular stacks combine proprioception with exteroception when available or operate purely proprioceptively in perception-degraded settings. MUSE~\cite{nistico2025muse}, building on~\cite{fink2020proprioceptive}, exemplifies this trend: a multi-sensor, real-time stack that can run proprioceptively when needed.

Among factor-graph approaches, Agrawal \textit{et al.}~\cite{agrawal2022proprioceptive} formulated proprioceptive state estimation using contact and kinematic factors, exploiting temporal consistency while maintaining real-time performance. In parallel, the Invariant Extended Kalman Filter (IEKF)~\cite{hartley2020contact} leverages Lie group structure so that the linearized error dynamics are state-independent, improving convergence and consistency under contact events; recent variants incorporate robust measurement costs and slip awareness~\cite{santana2024proprioceptive}. Smoothing methods extend this idea by optimizing over short windows with structure-aware residuals: the Invariant Smoother~\cite{yoon2024invariant} adapts IEKF group-affine residuals to fixed-lag optimization, yielding state-independent Jacobians and improved tolerance to dynamic contacts (e.g., slips), at higher computational cost.

\paragraph{\textbf{Contributions}}
In our study, we present, in a proprioceptive setting, an \emph{apples-to-apples} comparison of three representative state-of-the-art estimators for quadruped robots: MUSE~\cite{nistico2025muse} (proprioceptive mode), the IEKF~\cite{hartley2020contact}, and the Invariant Smoother (IS)~\cite{yoon2024invariant}, on a common real-world dataset~\cite{frey2026grandtour}. 
These methods were selected because they represent recent and high-performing approaches in proprioceptive state estimation, are publicly available through open-source implementations enabling reproducible evaluation, and cover complementary estimation paradigms, namely filtering, smoothing, and invariant-error estimation. This allows us to provide a fair and informative comparison of the trade-offs between accuracy, latency, and computational cost in deployable quadruped systems.

Following common trajectory evaluation practice~\cite{sturm2012benchmark}, we report: (i) long-term accuracy via Absolute Trajectory Error (ATE), (ii) short-term accuracy via translational and rotational Relative Pose Error (RPE), and (iii) computation time per update, highlighting latency differences between filtering and smoothing.
We also include a concise discussion of commonly adopted metrics, clarifying what each captures and when it is most informative for control and deployment. Additionally, we briefly discuss the convergence behavior of the three estimators under poor initialization of the orientation.
\looseness=-1

\paragraph{\textbf{Outline}}
\Cref{sec:methods} briefly reviews the three methodologies (modeling assumptions, measurement cues). \Cref{sec:results} presents results on a shared dataset with ATE/RPE and runtime, followed by a discussion of findings and metric interpretation. \Cref{sec:conclusions} summarizes takeaways for estimator selection and outlines limitations and future work. Finally, \Cref{sec:supp_material} describes the supplementary material.
\vspace{-0.35cm}

\section{Methods}
\label{sec:methods}

We compare (i) \emph{MUSE}~\cite{nistico2025muse} in proprioceptive mode, (ii) the IEKF~\cite{hartley2020contact}, and the fixed-lag IS~\cite{yoon2024invariant}. 
All estimate floating-base pose $(\mathbf{R},\mathbf{p})\!\in\!\mathrm{SO}(3)\times\mathbb{R}^3$, and body velocity $\mathbf{v}\!\in\!\mathbb{R}^3$, from IMU, leg kinematics, and contact states of quadruped robots. 
\looseness=-1

\subsection{MUSE (proprioceptive-only)}

MUSE is a modular, real-time fusion stack. In this paper we disable exteroceptive blocks and use only IMU, joint encoders, and contact/force sensing~\cite{nistico2025muse}.
The main components of MUSE comprise an Attitude Observer, a Leg Odometry module, and a Sensor Fusion algorithm that outputs the robot's pose and twist.

\subsubsection{\textbf{Attitude Observer}}
To estimate the robot's orientation, MUSE employs
a cascaded Nonlinear Observer (NLO), and an eXogeneous Kalman Filter (XKF) from~\cite{grip:ges,johansen:xkf}. 
The XKF linearizes about the NLO's globally stable estimate, yielding global stability with near-optimal Kalman filter's performance. 
The observer state is 
$\mathbf{x}_{\mathrm{t}} = [\mathbf{q}_\mathrm{t}^\top {\mathbf{b}_\mathrm{t}^\mathbf{\omega}}^\top]^\top \in \mathbb{R}^7$ with quaternion $\mathbf{q}_\mathrm{t} \in \mathbb{R}^4$ and gyroscope bias $\mathbf{b}^\mathbf{\omega}_\mathrm{t} \in \mathbb{R}^3$.
Measurements are
$\mathbf{z = [{f^b_\mathrm{t}}^\top {m^b_\mathrm{t}}^\top]^\top \in \mathbb{R}^6}$, 
where $\mathbf{f^b_\mathrm{t}}$ is the accelerometer specific force, and 
$\mathbf{m^b_\mathrm{t}}$, is the magnetometer reading when available. If the magnetometer is unavailable or unreliable, $\mathbf{m^b_\mathrm{t}}$ is obtained by rotating a fixed reference vector (constant in the navigation frame) into the body frame using the previous attitude estimate.
Full NLO/XKF equations are given in~\cite{fink2020proprioceptive}.

\subsubsection{\textbf{Leg Odometry}}
Leg odometry measures the linear velocity of the floating base from the forward kinematics of legs in stable contact. The measured base velocity is
\begin{equation}
        \mathbf{v^b_\mathrm{t}} = \frac{1}{n_s}\sum_{\ell\in\mathbb{L}} \mathbf{v^b_\mathrm{t_\ell}},  
        \quad
        \mathbf{v^b_\mathrm{t_\ell}}=-\alpha_{\ell}\Big(\mathbf{J}_{\ell}(\mathbf{q}_{\ell})\,\dot{\mathbf{q}}_{\ell} \boldsymbol{\omega}^{b}\times \mathbf{d}^{b}_{\ell}\Big)
\end{equation}
where $\mathbb{L}$ is the set of legs, $\alpha_{\ell}\!\in\!\{0,1\}$ indicates contact state, and $n_s = \sum_{\ell\in\mathbb{L}} \alpha_{\ell}$ is the number of stance legs. The contribution of leg $\ell$ is $\mathbf{v^b_\mathrm{t_\ell}}$,
with $\mathbf{J}_{\ell}(\mathbf{q}_{\ell})$ the leg Jacobian, $\mathbf{q}_{\ell}$ joint position, $\dot{\mathbf{q}}_{\ell}$ joint velocity, $\boldsymbol{\omega}^{b}$ base angular velocity, $\mathbf{d}^{b}_{\ell}$ foot position in the base frame, and $\mathbf{v}^{b}_{\ell}$ the foot velocity in the base frame. 

\subsubsection{\textbf{Sensor Fusion}}
The IMU and leg odometry measurements are fused in a Kalman Filter, 
with dynamics as in~\cite{fink2020proprioceptive}.
The state $\mathbf{x_\mathrm{t} = [{p_\mathrm{t}}^\top {v_\mathrm{t}}^\top]^\top \in \mathbb{R}^6}$ is the position and 
velocity of the base, the input $\mathbf{u_\mathrm{t} = (R_\mathrm{t} f_\mathrm{t}^b - g) \in \mathbb{R}^3}$ is the base acceleration, and the measurement is the leg odometry velocity $z_\mathrm{t} = [\mathbf{R_\mathrm{t} v_\mathrm{t}^b}]^\top \in \mathbb{R}^3$. The filter uses
$\mathbf{K \in \mathbb{R}^{6 \times 3}}$ (Kalman gain), $\mathbf{P \in \mathbb{R}^{6 \times 6}}$ (covariance matrix), $\mathbf{Q \in \mathbb{R}^{6 \times 6}}$ (process noise), and $\mathbf{R \in \mathbb{R}^{3 \times 3}}$ (measurement noise covariance matrix).

\subsection{IEKF}
The theory in this section follows~\cite{hartley2020contact}. The IEKF is posed on a group-affine system, so the right-invariant error is state-independent. Propagation integrates IMU data; the update assumes stable contact (zero foot velocity). The continuous-time dynamics with noise and bias are:
\begin{align}
        \frac{\mathrm{d}}{\mathrm{dt}}\mathbf{R}_{\mathrm{t}} &= \mathbf{R}_{\mathrm{t}}(\mathbf{\tilde{\omega}}_{\mathrm{t}}-\mathbf{b}^\mathbf{\omega}_{\mathrm{t}} - \mathbf{w}^{\mathbf{\omega}}_\mathrm{t})^{\wedge}, \quad  
        \frac{\mathrm{d}}{\mathrm{dt}}\mathbf{p}_{\mathrm{t}} = \mathbf{v}_{\mathrm{t}}
        \label{eq:dyn1}
        \\
        \frac{\mathrm{d}}{\mathrm{dt}}\mathbf{v}_{\mathrm{t}} &= \mathbf{R}_{\mathrm{t}}(\mathbf{\tilde{a}}_{\mathrm{t}} - \mathbf{b}^{\mathbf{a}}_{\mathrm{t}} - \mathbf{w}^\mathbf{a}_\mathrm{t}) + \mathbf{g},\quad \frac{\mathrm{d}}{\mathrm{dt}}\mathbf{d}_{\mathrm{t}}  = \mathbf{R}_{\mathrm{t}} \mathbf{w}^\mathbf{d}_{\mathrm{t}}\label{eq:dyn2} \\
        \frac{\mathrm{d}}{\mathrm{dt}}\mathbf{b}^{\mathbf{\omega}}_{\mathrm{t}} &= \mathbf{w}^{\mathbf{b}^\mathbf{\omega}}_\mathrm{t}, \quad \frac{\mathrm{d}}{\mathrm{dt}}\mathbf{b}^{\mathbf{a}}_{\mathrm{t}} = \mathbf{w}^{\mathbf{b}^\mathbf{a}}_\mathrm{t} \label{eq:dyn3}  
\end{align} 
Here $\mathbf{g}$ is gravity and all $\mathbf{w}^{(\cdot)}$ are zero-mean Gaussian.

The discrete, noise-corrupted log-linear propagation is
\begin{equation}
\begin{bmatrix}\xi^{\mathrm{r}}_\mathrm{t+1}\\
\zeta^{\mathrm{r}}_\mathrm{t+1}
\end{bmatrix}
= (\mathbb{I}_{18}+\mathbf{A}_{\mathrm{t}}\Delta{t}) 
\begin{bmatrix}
\xi^{\mathrm{r}}_\mathrm{t}\\
\zeta^{\mathrm{r}}_\mathrm{t}
\end{bmatrix}
+\mathbf{B}_{\mathrm{t}}\bar{\mathbf{w}}_\mathrm{t},
\mathbf{B}_{\mathrm{t}}=\Delta{t}
\begin{bmatrix}
\text{Ad}_{\bar{\mathbf{X}}_\mathrm{t}} \quad \mathbf{0}_{12,6} \\
\mathbf{0}_{6,12} \quad  \mathbb{I}_{6}
\end{bmatrix}
    \label{eq:log_linear_prop}
\end{equation}
\begin{equation}
\mathbf{A}_{\mathrm{t}}=\begin{bmatrix}\mathbf{0} & \mathbf{0}_{3,3} & \mathbf{0}_{3,3} & \mathbf{0}_{3,3} & -\overline{\mathbf{R}}_\mathrm{t} & \mathbf{0}_{3,3} \\ (\mathbf{g})^{\wedge} & \mathbf{0}_{3,3} & \mathbf{0}_{3,3} & \mathbf{0}_{3,3} & -\left(\overline{\mathbf{v}}_\mathrm{t}\right)^{\wedge} \overline{\mathbf{R}}_\mathrm{t} & -\overline{\mathbf{R}}_\mathrm{t} \\ \mathbf{0}_{3,3} & \mathbb{I}_{3} & \mathbf{0}_{3,3} & \mathbf{0}_{3,3} & -\left(\overline{\mathbf{p}}_\mathrm{t}\right)^{\wedge} \overline{\mathbf{R}}_\mathrm{t} & \mathbf{0}_{3,3} \\ \mathbf{0}_{3,3} & \mathbf{0}_{3,3} & \mathbf{0}_{3,3} & \mathbf{0}_{3,3} & -\left(\overline{\mathbf{d}}_\mathrm{t}\right)^{\wedge} \overline{\mathbf{R}}_\mathrm{t} & \mathbf{0}_{3,3} \\ \mathbf{0}_{6,3} & \mathbf{0}_{6,3} & \mathbf{0}_{6,3} & \mathbf{0}_{6,3} & \mathbf{0}_{6,3} & \mathbf{0}_{6,3} 
\end{bmatrix},
    \label{eq:A_matrix_IS}
\end{equation}
with time-step $\Delta{t}$, and noise vector $\bar{\mathbf{w}}_\mathrm{t} = [(\mathbf{w}^{\omega}_\mathrm{t})^{\top}$,$ (\mathbf{w}^{\mathbf{a}}_\mathrm{t})^{\top}$,$ (\mathbf{w}^{\mathbf{a}}_\mathrm{t}\Delta{t})^{\top}$,$ (\mathbf{w}^{\mathbf{d}}_\mathrm{t})^{\top}$,$(\mathbf{w}^{b^{\omega}}_{t})^{\top}$,$(\mathbf{w}^{b^{a}}_{t})^{\top}]^{\top}$.

For the observation, the contact measurement is written in right-invariant, group-affine form
$\mathbf{y}_\mathrm{t} = \mathbf{X}_\mathrm{t}^{-1} \mathbf{s} + \mathbf{w}^{\text{obs}}_\mathrm{t} $, with forward kinematics $\mathrm{fk}(\mathbf{\tilde{q}}_\mathrm{t}) = \mathbf{R}^\top_\mathrm{t}(\mathbf{d}_\mathrm{t} -\mathbf{p}_\mathrm{t})+\mathbf{J}_\mathrm{p}(\mathbf{\tilde{q}}_\mathrm{t})\mathbf{w}^{q}_\mathrm{t}$. As in~\cite{hartley2020contact}, the right-invariant measurement is
\begin{equation}
        \mathbf{Y}^{\mathrm{kin}}_{\mathrm{t}} = \mathbf{X}^{-1}_{\mathrm{t}} \mathbf{b}^{\mathrm{kin}}_{\mathrm{t}} + \mathbf{V}^{\mathrm{kin}}_{\mathrm{t}},
        \label{eq:kine_meas}
\end{equation}
where $\mathbf{Y}^{\mathrm{kin}}_{\mathrm{t}} = [\mathrm{fk}(\mathbf{\tilde{q}}_\mathrm{t}) ~ 0 ~ 1 ~ -1]^\top$ is the vector of kinematic observation, $\mathbf{X}^{-1}_{\mathrm{t}}$ is the inverse of the state matrix~$\mathbf{X}_{\mathrm{t}}$,  
${\mathbf{b}^{\mathrm{kin}}_{\mathrm{t}}} = [\mathbf{0}_{3,1} ~ 0 ~ 1 ~ -1]^\top $ is a constant vector, while ${\mathbf{V}^{\mathrm{kin}}_{\mathrm{t}}} = [\mathbf{J_p}(\mathbf{\tilde{q}_t}) \mathbf{w_t^q} ~ 0 ~ 0 ~ 0]^\top$     is the Gaussian noise vector of the observation model.
Given~\eqref{eq:kine_meas}, the IEKF update is
\[\bar{\mathbf{X}}^{+}_{t}=\mathrm{Exp}(\mathbf{K}_{\mathrm{t}}\Pi\bar{\mathbf{X}}^{-}_{\mathrm{t}}\mathbf{Y}^{\mathrm{kin}}_{\mathrm{t}})\bar{\mathbf{X}}^{-}_{\mathrm{t}}\] \[ \mathbf{P}^{+}_{t}=(\mathbb{I}-\mathbf{K}_{\mathrm{t}}\mathbf{H}_{\mathrm{t}})\mathbf{P}_{\mathrm{t}}(\mathbb{I}-\mathbf{K}_{\mathrm{t}}\mathbf{H}_{\mathrm{t}})^{\top}+\mathbf{K}_{\mathrm{t}}\bar{\mathbf{N}}_{\mathrm{t}}\mathbf{K}^{\top}_{\mathrm{t}}\]
with $\Pi\triangleq\begin{bmatrix}\mathbb{I} \quad\mathbf{0}_{3,3}\end{bmatrix}$, $\mathbf{K}_{\mathrm{t}}$ is the Kalman gain matrix computed as $\mathbf{K}_{\mathrm{t}} =\mathbf{P}_{\mathrm{t}}\mathbf{H}_{\mathrm{t}}^{\top}(\mathbf{H}_{\mathrm{t}}\mathbf{P}_{\mathrm{t}}\mathbf{H}^{\top}_{\mathrm{t}}+\bar{\mathbf{N}}_{\mathrm{t}})^{-1}$. The matrices $\mathbf{H}_{\mathrm{t}}$ and $\bar{\mathbf{N}}_{\mathrm{t}}$ are given by $\mathbf{H}_{\mathrm{t}}=\begin{bmatrix}
\mathbf{0}_{3,3} \quad \mathbf{0}_{3,3} \quad -\mathbb{I} \quad \mathbb{I}\end{bmatrix}$ and $\bar{\mathbf{N}}_{\mathrm{t}}=\mathbf{R}^{-}_{\mathrm{t}}\mathbf{J_{p}}(\mathbf{\tilde{q}_\mathrm{t}})\text{Cov}(\mathbf{w}^{\mathbf{q}}_{\mathrm{t}})\mathbf{J_{p}}^{\top}(\mathbf{\tilde{q}_\mathrm{t}})(\mathbf{R}^{-}_{\mathrm{t}})^{\top}.$

\subsection{Invariant Smoother}
Unlike the IEKF, which updates only the most recent states, the IS in~\cite{yoon2024invariant} seeks a Maximum A Posteriori (MAP) estimate over a fixed window of size $n$ using states $\mathbf{X}_\mathrm{0:n}$ and measurements $\mathbf{Z}_\mathrm{0:n}$. MAP is formulated as a nonlinear least-squares problem:
\begin{equation}
\begin{aligned}
    &\operatorname*{\mathbf{e}^{*}}_{0:\mathrm{n}} 
    = \operatorname*{argmin}_{\mathbf{e}_{0:\mathrm{n}}} ({\| \mathbf{r}_{\text{pri}} - \mathbf{J}_{\text{pri}}\mathbf{e}_{0:\mathrm{n}} \|^2_{\Sigma_{\text{pri}}}} 
    \\ &+ {\sum_{{{\mathrm{t}}}=0}^{\mathrm{n}-1} \| \mathbf{r}^{{\mathrm{t}}}_{\text{pro}}-\mathbf{J}^{{\mathrm{t}}}_{\text{pro}}\mathbf{e}_{0:n}\|^2_{\Sigma_{\text{pro}}}}  
    + {\sum_{{{\mathrm{t}}}=0}^{\mathrm{n}} \| \mathbf{r}^{{\mathrm{t}}}_{\text{o}}-\mathbf{J}^{{\mathrm{t}}}_{\text{o}}\mathbf{e}_{0:n} \|^2_{\Sigma_{\text{o}}}}),
    \end{aligned}
\label{eq:is_cost}
\end{equation}
where $\mathrm{n}$ is the window size (WS), $\mathbf{r}_{\cdot}$ are the residuals, $\Sigma_{\cdot}$ the covariances, and $\mathbf{J}_{\cdot}$ the Jacobians at the operating points $\bar{\mathbf{X}}_{\mathrm{t}}$ and $\bar{\mathbf{x}}_{\mathrm{t}}$ for the \textit{prior}, \textit{propagation}, and \textit{observation} distributions, respectively. The prior anchors the window; marginalization keeps its size fixed~\cite{yoon2024invariant}.

\paragraph{\textbf{Propagation Cost}} From the log-linear property we get the residual~\cite{yoon2024invariant}: 
\begin{equation}
\mathbf{r}^{\mathrm{t}}_{\text{pro}}=\begin{bmatrix}
        \mathrm{Log}(f^{\mathrm{d}}_{M}(\bar{\mathbf{X}}_{{\mathrm{t}}})\bar{\mathbf{X}}^{-1}_{t+1}) \\ f^{\mathrm{d}}_{v}(\bar{\mathbf{x}}_{{\mathrm{t}}})-\bar{\mathbf{x}}_{{\mathrm{t}}+1}
    \end{bmatrix},
    \Sigma_{\text{pro}}=\mathbf{B}_{\mathrm{t}}\mathrm{Cov}(\bar{\mathbf{w}}_{\mathrm{t}})\mathbf{B}^{\top}_{\mathrm{t}},
\end{equation}
with Jacobians $\mathbf{J}^{{\mathrm{t}}+1}_{\mathrm{pro}}=\mathbb{I}_{18}+\mathbf{A}_{\mathrm{t}}\Delta{t}$ and $\mathbf{J}^{{\mathrm{t}}}_{\mathrm{pro}}=\mathbb{I}_{18}$. The discrete dynamics are obtained by discretizing~\eqref{eq:dyn1}-\eqref{eq:dyn3}:
\begin{equation}
    f^{\mathrm{d}}_{M}({\mathbf{X}}_{{\mathrm{t}}})=\begin{bmatrix}
        \mathbf{R}^{\mathrm{d}}_{{\mathrm{t}}} \quad &\mathbf{v}^{\mathrm{d}}_{{\mathrm{t}}} \quad
        &\mathbf{p}^{\mathrm{d}}_{{\mathrm{t}}} \quad
        &\mathbf{d}^{\mathrm{d}}_{{\mathrm{t}}} \\ \mathbf{0}_{3,3} 
        \quad 
        &\mathbf{}
        \quad
        &\mathbb{I}_{3}
        \quad
        &\mathbf{}
    \end{bmatrix}, 
    f^{\mathrm{d}}_{v}(\mathbf{x}_{{\mathrm{t}}})=\begin{bmatrix}
    \mathbf{b}^{\omega}_{{\mathrm{t}}} \\
    \mathbf{b}^{\mathbf{a}}_{{\mathrm{t}}}
    \end{bmatrix},
\end{equation}
where $\mathbf{R}^{\mathrm{d}}_{{\mathrm{t}}}=\mathbf{R}_{{\mathrm{t}}}\mathrm{Exp}((\bar{\omega}_{{\mathrm{t}}}-\mathbf{b}^{\omega}_{{\mathrm{t}}})\Delta{t})$, $\mathbf{v}^{\mathrm{d}}_{{\mathrm{t}}}=\mathbf{v}_{{\mathrm{t}}}+\mathbf{R}_{{\mathrm{t}}}(\bar{\mathbf{a}}_{{\mathrm{t}}}-\mathbf{b}^{\mathbf{a}}_{{\mathrm{t}}})\Delta{t}+\mathbf{g}\Delta{t}$, and $\mathbf{p}^{\mathrm{d}}_{{\mathrm{t}}}=\mathbf{p}_{{\mathrm{t}}}+\mathbf{v}_{{\mathrm{t}}}\Delta{t}+\frac{1}{2}\mathbf{R}_{{\mathrm{t}}}(\bar{\mathbf{a}}_{{\mathrm{t}}}-\mathbf{b}^{\mathbf{a}}_{{\mathrm{t}}})(\Delta{t})^{2}+\frac{1}{2}\mathbf{g}(\Delta{t})^{2}$.

\paragraph{\textbf{Observation Cost}} Using the right-invariant kinematics observation of~\eqref{eq:kine_meas} as in~\cite{yoon2024invariant}:
\begin{equation}
\mathbf{r}^{{\mathrm{t}}}_{\mathrm{o}}= \mathbf{X}_{{\mathrm{t}}}\mathbf{Y}^{\mathrm{kin}}_\mathrm{t} - \mathbf{b}^{\mathrm{kin}}_{{\mathrm{t}}}
\end{equation}
\begin{equation}
\mathbf{J_{o}}^\mathrm{t} = [(\mathbf{b}^{\mathrm{kin}}_{\mathrm{t}})^\odot \hspace{0.2cm} \mathbf{0}_{6,6}]
,\quad {\Sigma}^{{\mathrm{t}}}_{\mathbf{o}} = \bar{\mathbf{X}}_{{\mathrm{t}}} {\Sigma_{\mathrm{kin}}(w^q)_{\mathrm{t}}} \bar{\mathbf{X}}^\top_{{\mathrm{t}}},
\end{equation}
where the $\odot$ operator is defined in~\cite{yoon2024invariant}. Finally, solving \eqref{eq:is_cost} (e.g., Gauss-Newton) yields increments that update the operating points each iteration:
$\mathbf{X}^{*}_{\mathrm{t}}\leftarrow\mathrm{Exp}(\xi^{r*}_{\mathrm{t}})\bar{\mathbf{X}}_{\mathrm{t}}, \mathbf{x}^{*}_{\mathrm{t}}\leftarrow\bar{\mathbf{x}}_{i}+\zeta^{r*}_{i}$.

\begin{figure*}[!t]
    \centering
    \begin{subfigure}[b]{0.27\textwidth}
        \includegraphics[width=\textwidth]{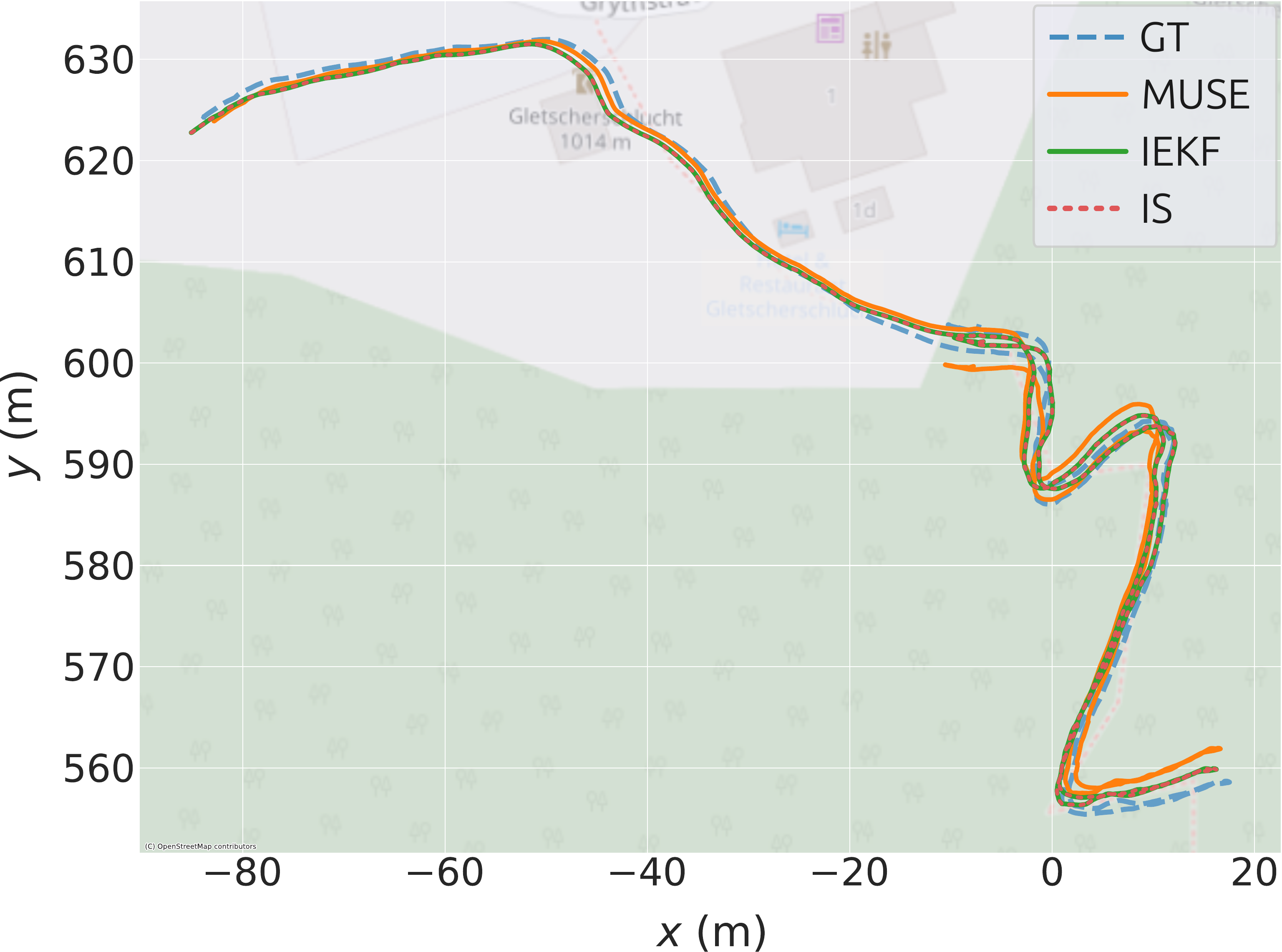}
        \caption{GT vs. estimated trajectories.}
        \label{fig:traj_comparison}
    \end{subfigure}
    \hfill
    \begin{subfigure}[b]{0.29\textwidth}
        \includegraphics[width=\textwidth]{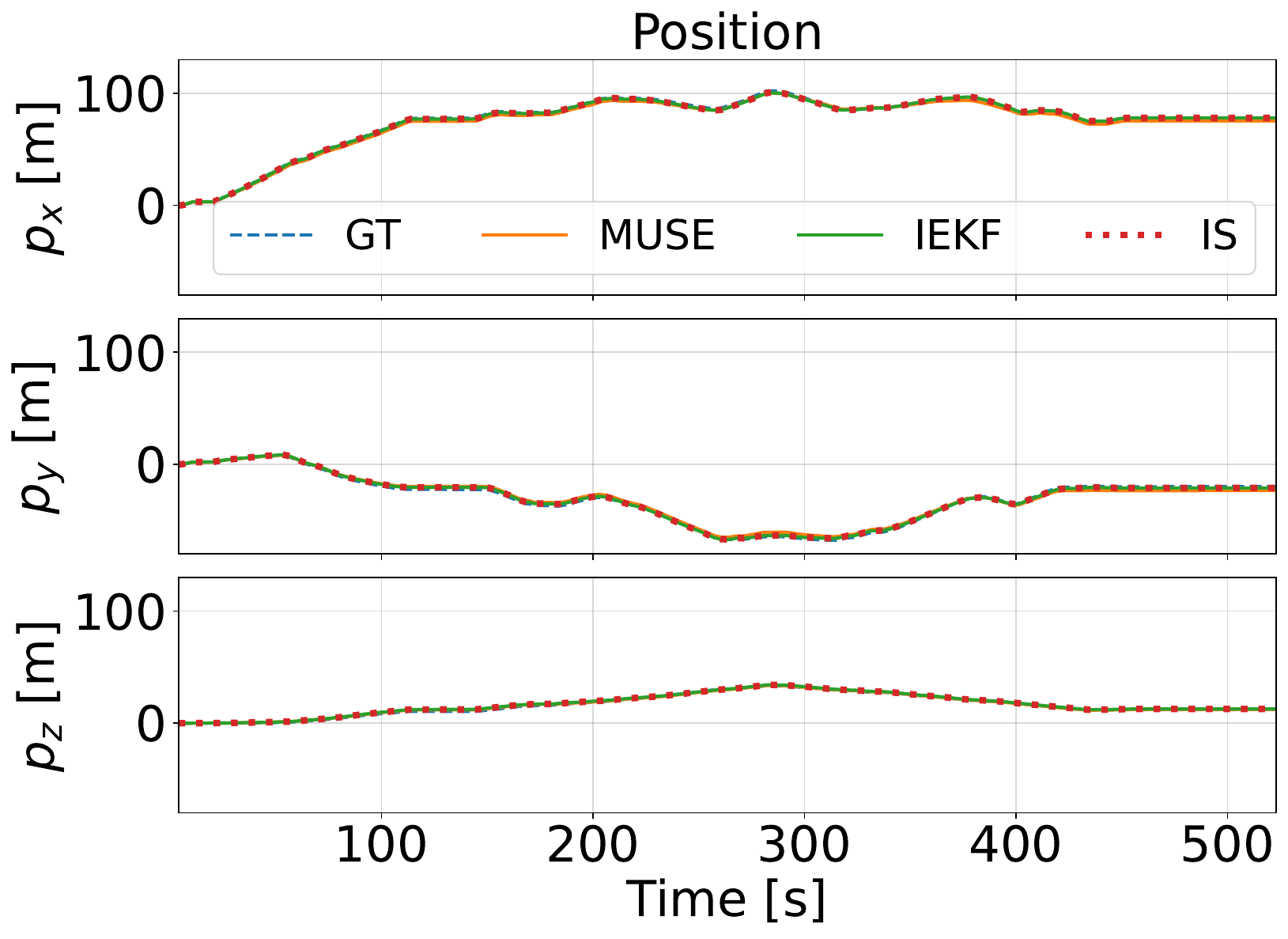}
        \caption{GT vs. estimated positions.}
        \label{fig:pos_comparison}
    \end{subfigure}
    \hfill
    \begin{subfigure}[b]{0.29\textwidth}
        \includegraphics[width=\textwidth]{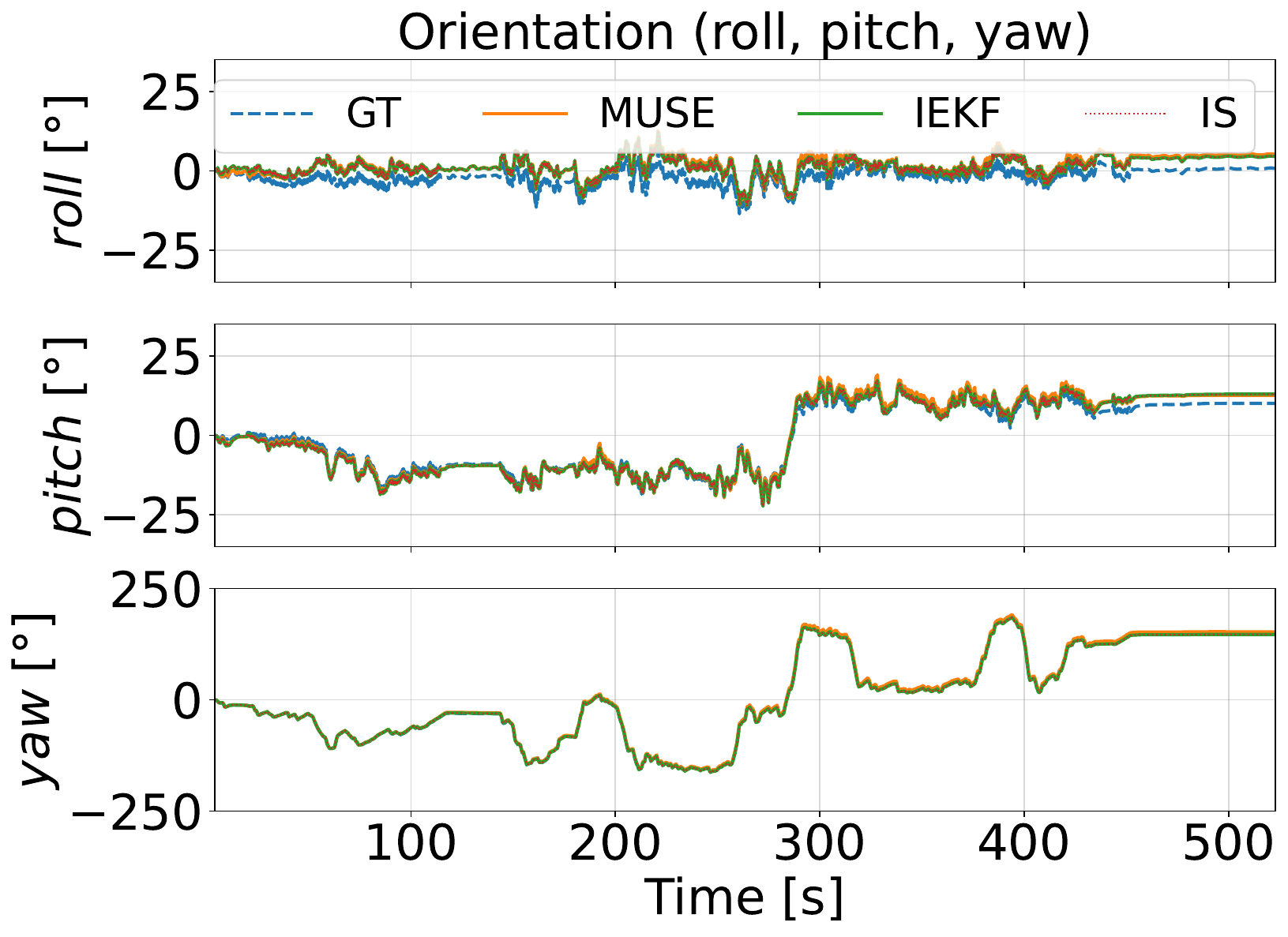}
        \caption{GT vs. estimated orientations.}
        \label{fig:rpe_comparison}
    \end{subfigure}

    \caption{ Comparison of the ground truth (GT) trajectory, position and orientation vs. the estimates obtained with MUSE, IEKF, and IS, on the CYN-1 (Grindelwald Canyon) sequence of the Grand Tour Dataset~\cite{frey2026grandtour}. 
    }
    \label{fig:pose_comparison}
\end{figure*}

\begin{figure*}[!t]
    \centering
    \begin{subfigure}[b]{0.3\textwidth}
        \vspace{-0.35cm}
        \includegraphics[width=\textwidth]{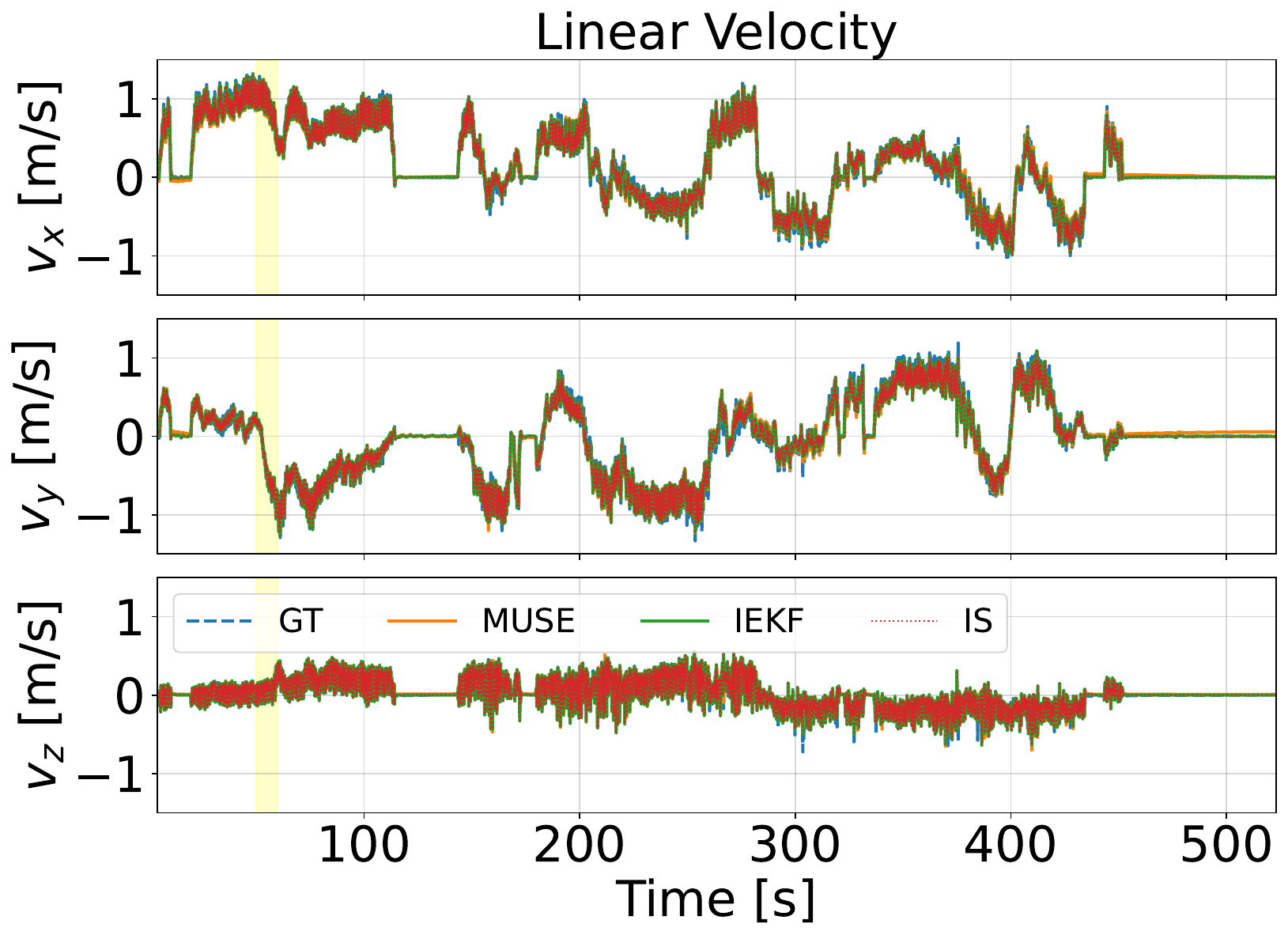}
        \caption{GT vs. estimated velocities.}
        \label{fig:vel_comparison}
    \end{subfigure}
    \hfill
    \begin{subfigure}[b]{0.3\textwidth}
        \vspace{-0.35cm}
        \includegraphics[width=\textwidth]{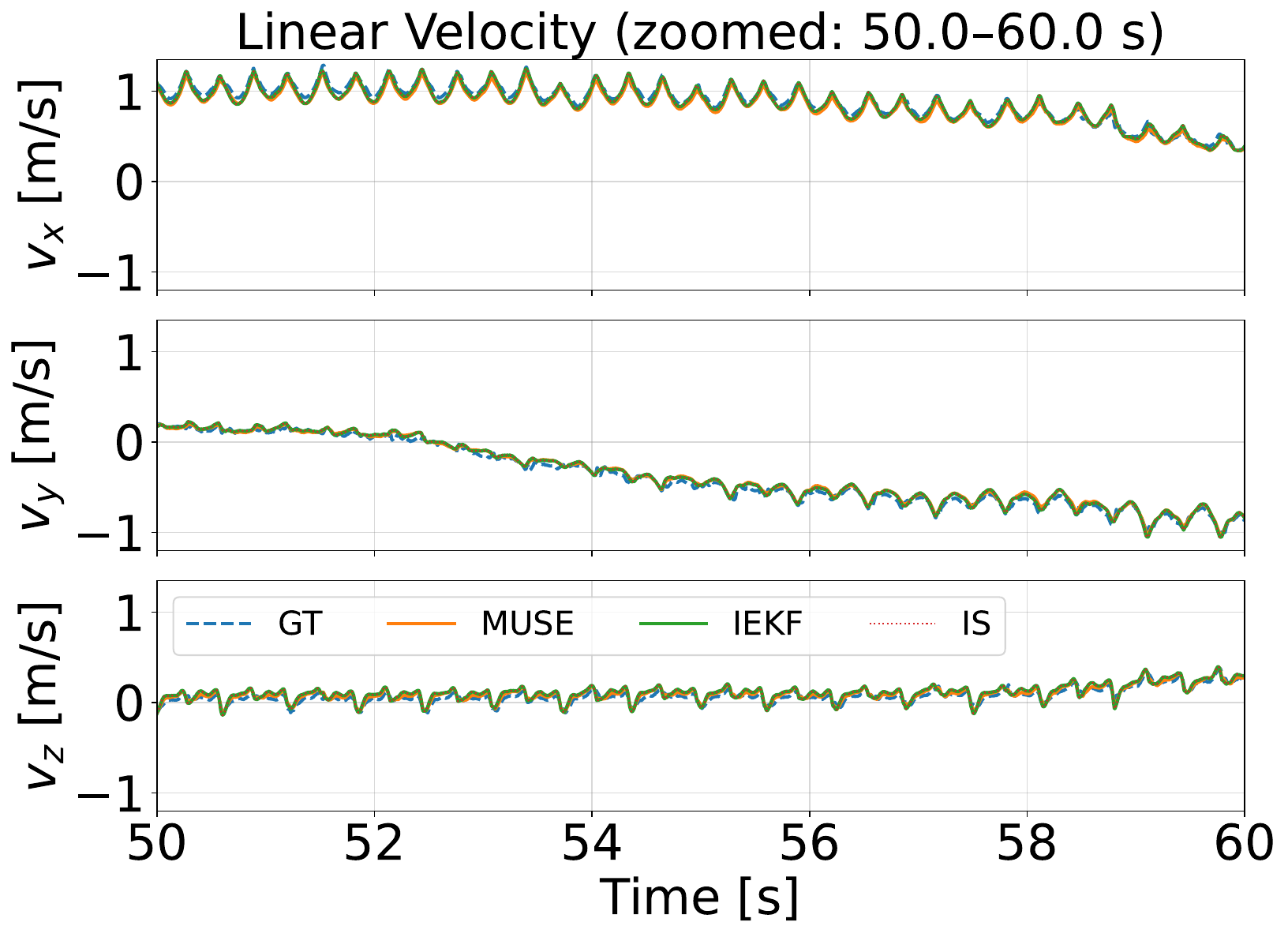}
        \caption{GT vs. estimated velocities (zoomed).}
        \label{fig:vel_zoomed_comparison}
    \end{subfigure}
    \hfill
    \begin{subfigure}[b]{0.3\textwidth}
        \vspace{-0.25cm}
        \begin{tabular}{c}
            \includegraphics[width=0.9\textwidth]{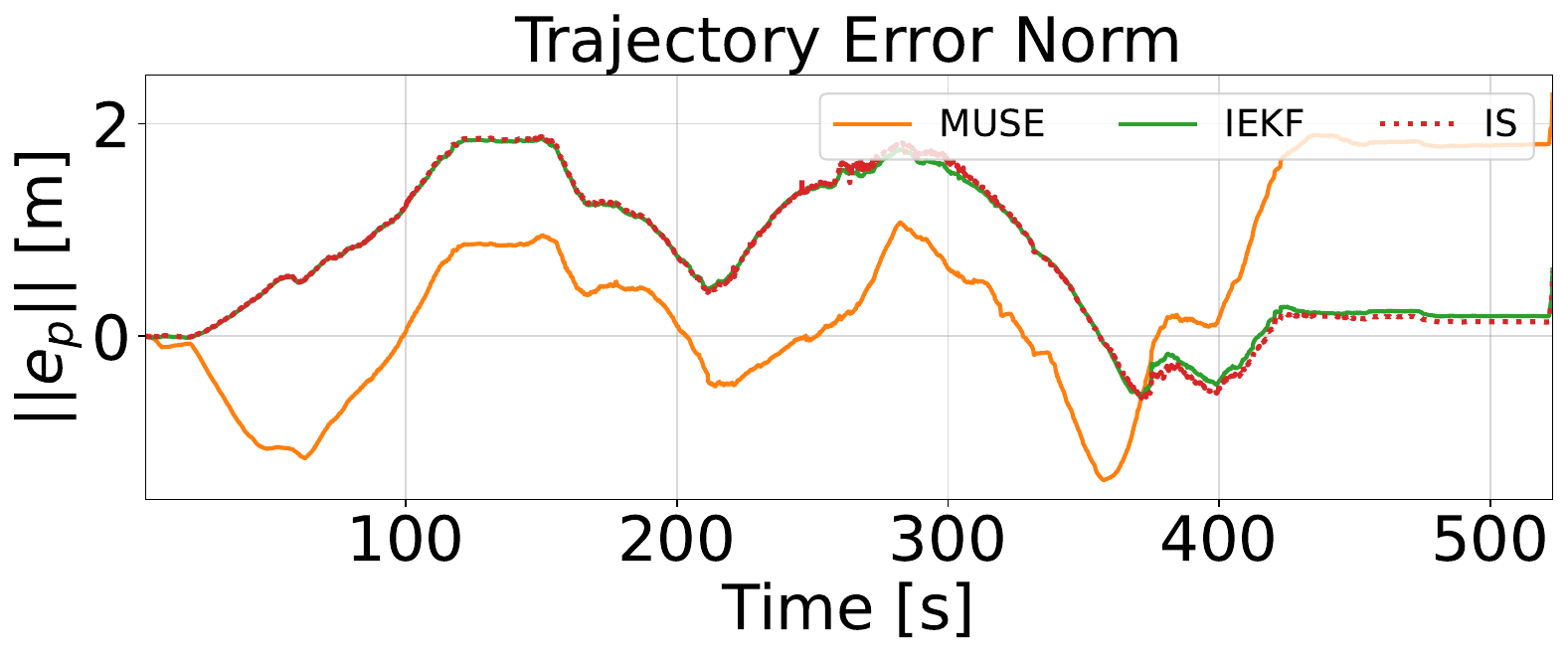} \\
            \vspace{-0.55cm}
            \includegraphics[width=0.9\textwidth]{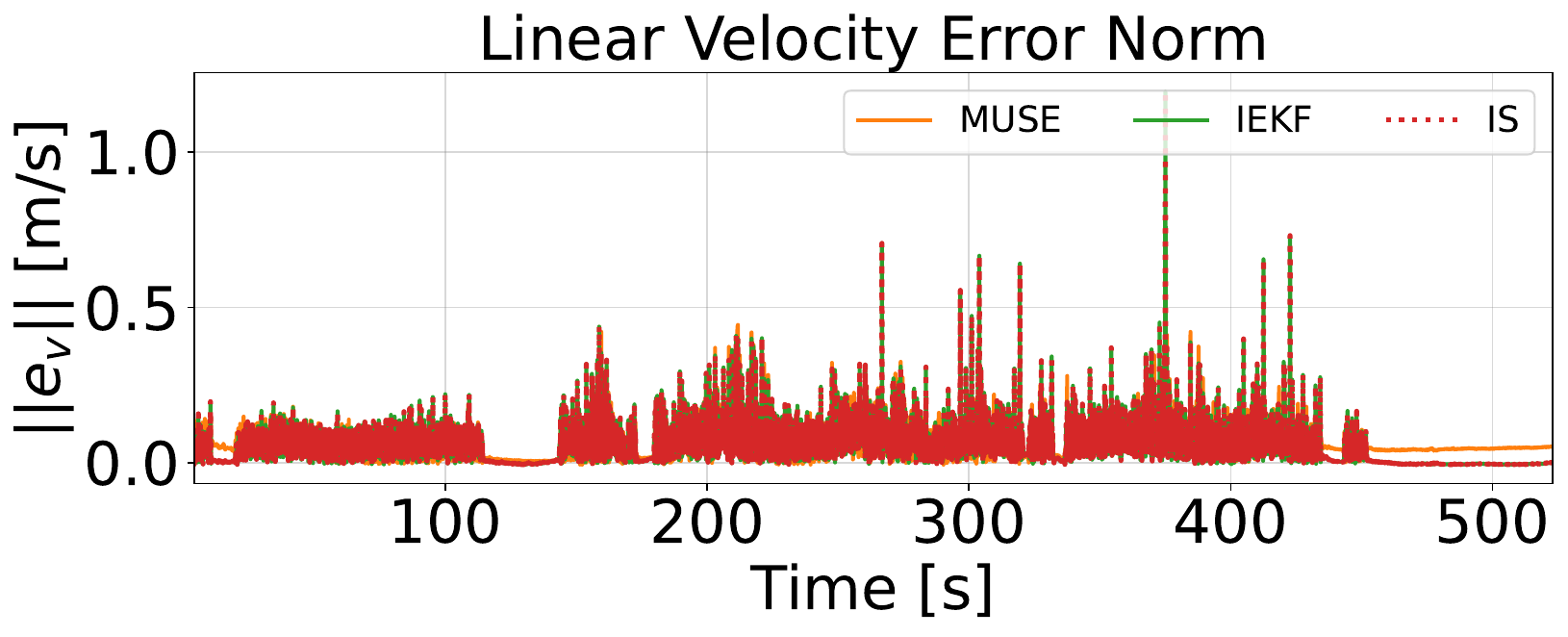}
        \end{tabular}
        \caption{Trajectory and velocity errors over time.}
        \label{fig:errors_comparison}
    \end{subfigure}

    \caption{ 
    (a) Comparison of ground-truth (GT) and estimated velocities from MUSE, IEKF, and IS. (b) Zoomed-in view of the highlighted interval. (c) Error metrics over time, with trajectory error norm (top) and velocity error norm (bottom).
    \vspace{-0.55cm}
    }.

    \label{fig:velocity_comparison}
\end{figure*}

\section{Results and Discussion}
\label{sec:results}

\subsection{Evaluation Metrics}
We evaluate performance using two standard Root Mean Square Error (RMSE)-based metrics. The Absolute Trajectory Error (ATE) measures global pose accuracy by aligning the estimated trajectory to the ground-truth and computing the RMSE over the full path (captures accumulated drift). The Relative Pose Error (RPE) measures short-horizon accuracy by computing windowed, or frame-to-frame motion differences and taking their RMSE (captures local consistency). Both metrics are computed with the \texttt{evo} Python package~\cite{grupp2017evo}.
We also report velocity RMSE, by comparing the estimated and GT trajectories at each timestamp. This metric captures the accuracy of the estimated velocity profiles, which is important for control and planning applications.
\looseness=-1

Here, we report IS results using a WS of 3, which balances accuracy and runtime.  We analyze the effect of window size on accuracy and runtime in \Cref{sec:computation_times}.

\subsubsection{\textbf{Absolute Trajectory Error (ATE)}}
Each estimated trajectory is aligned to the ground truth with SE(3) Umeyama \cite{grupp2017evo}, and the RMSE is computed using \texttt{evo} for interpolation (linear in position, SLERP in orientation). Results are reported in \Cref{tab:ate_rpe}.
All three pipelines achieve relatively similar ATE values. The difference between the best (IS) and worst (MUSE) result is approximately 0.91 m, which corresponds to about $0.3\%$ of the total 300~m trajectory length.
Because SE(3) alignment removes constant pose bias, the residual ATE mainly reflects accumulated drift and modeling errors. IEKF and IS achieve lower ATE values than MUSE, which is consistent with their group-affine modeling and, in the case of IS, the additional smoothing regularization. 

3D overlays (\Cref{fig:traj_comparison}) and per-axis (x-y-z) plots (\Cref{fig:pos_comparison,fig:rpe_comparison}) show trajectories that remain broadly aligned over the full sequence, although MUSE exhibits a larger end-point deviation. This interpretation is further supported by the trajectory error norm in \Cref{fig:errors_comparison}, which remains bounded over time and indicates moderate differences among the estimators.
\looseness=-1

\begin{table}[!b]
        \begin{center}
            \caption{\textbf{Grand Tour Dataset}: ATE, ATE$_{\text{vel}}$, and RPE  ($\sim$ 296 m trajectory) errors. The best result per row is highlighted in \textbf{bold}, the second-best in \underline{underline}.}
            \label{tab:ate_rpe}
            \begin{tabular}{c c c c}
                \toprule
                RMSE & MUSE & IEKF & IS  \\
                \midrule
                \rowcolor{gray!15}
                ATE [m] & 2.269461 & \underline{1.405668} & \textbf{1.363114} \\
                ATE$_{\text{vel}}$  [m/s] & 0.876147 & \textbf{0.869383} & \underline{0.869481} \\
                \rowcolor{gray!15}
                RPE ($\Delta=$ 1 meter) [m] & 0.072223 &  \underline{0.043187} &  \textbf{0.042526} \\
                RPE ($\Delta=$ 1 frame) [m] &\underline{0.000670} & \textbf{0.000544} & 0.001965 \\
                \rowcolor{gray!15}
                RPE ($\Delta=$ 1 meter) [$\circ$] & 0.578469 & \underline{0.568987} & \textbf{0.565644}  \\
                RPE ($\Delta=$ 1 frame) [$\circ$] & \textbf{0.002605} & \underline{0.002611} & 0.002614 \\
                \bottomrule
            \end{tabular}
            \end{center}
        \end{table}

\subsubsection{\textbf{Relative Pose Error (RPE)}}
\label{sec:rpe_discussion}
We report translational and rotational RPEs in \Cref{tab:ate_rpe} using \texttt{evo} under two windowing schemes:
(i) a spatial window with $\Delta=1$ m, evaluating motion over segments of approximately one meter of reference path length (probing meter-scale drift); and
(ii) a temporal window with $\Delta=1$ frame, corresponding to one sample in the synchronized logs (probing frame-to-frame consistency).

At $\Delta=1~\text{m}$, IEKF and IS achieve lower translational RPE than MUSE (approximately $40\%$ reduction). At $\Delta=1$ frame, IEKF and MUSE exhibit lower RPE than IS (approximately $72\%$ and $66\%$ reduction).
The stronger separation under the meter-based window suggests that small systematic errors (e.g., micro-slips, encoder biases, or residual IMU-kinematics coupling) accumulate over the multiple updates required to traverse one meter. Differences in filtering and contact handling likely contribute at this scale. In contrast, the frame-based metric is more sensitive to measurement jitter and estimator bandwidth; here, filters' higher-rate fusion appears to better suppress high-frequency noise.
Similar consideration applies to Rotational RPE, which is comparable across methods, with minor differences possibly linked to attitude-estimation stability properties.

\subsubsection{\textbf{Linear Velocity}} The RMSE of the velocity estimates, reported in \Cref{tab:ate_rpe} and visualized in \Cref{fig:vel_comparison,fig:vel_zoomed_comparison,fig:errors_comparison}, shows that the differences are small, and all three methods provide accurate velocity profiles that closely match the ground truth.

\subsection{Runtime and Computational Load}
\label{sec:computation_times}
\begin{figure}
    \centering
    \includegraphics[width=0.42\textwidth]{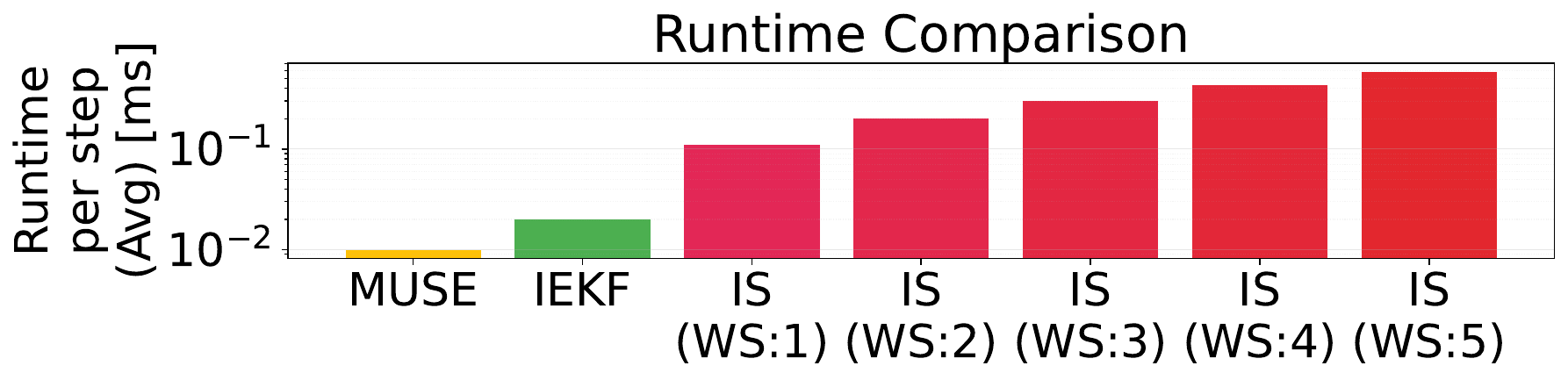}
    \caption{Computation time per iteration for MUSE, IEKF, and IS with different window sizes (WS). A log scale is used for the y-axis for a clearer visualization.}
    \label{fig:computation_time}
    \vspace{-0.45cm}
\end{figure}

To assess real-time suitability, we measured per-iteration runtime for each estimator (current implementations), summarized in \Cref{fig:computation_time}. MUSE is the fastest ($\approx$0.012 ms), followed by IEKF ($\approx$0.02 ms). In contrast, IS exposes an explicit accuracy--latency trade-off via its window size (WS): from $\approx$0.110 ms at $\mathrm{WS}=1$ up to $\approx$0.583 ms at $\mathrm{WS}=5$.

A practical implication is that fixed-lag smoothing introduces algorithmic latency proportional to the lag: estimates may be delayed depending on whether the pipeline outputs the lagged state or a real-time approximation. This delay is negligible for small WS but becomes significant as WS increases.

\subsection{Convergence with large initial conditions}
\begin{figure}
    \centering
    \includegraphics[width=0.4\textwidth]{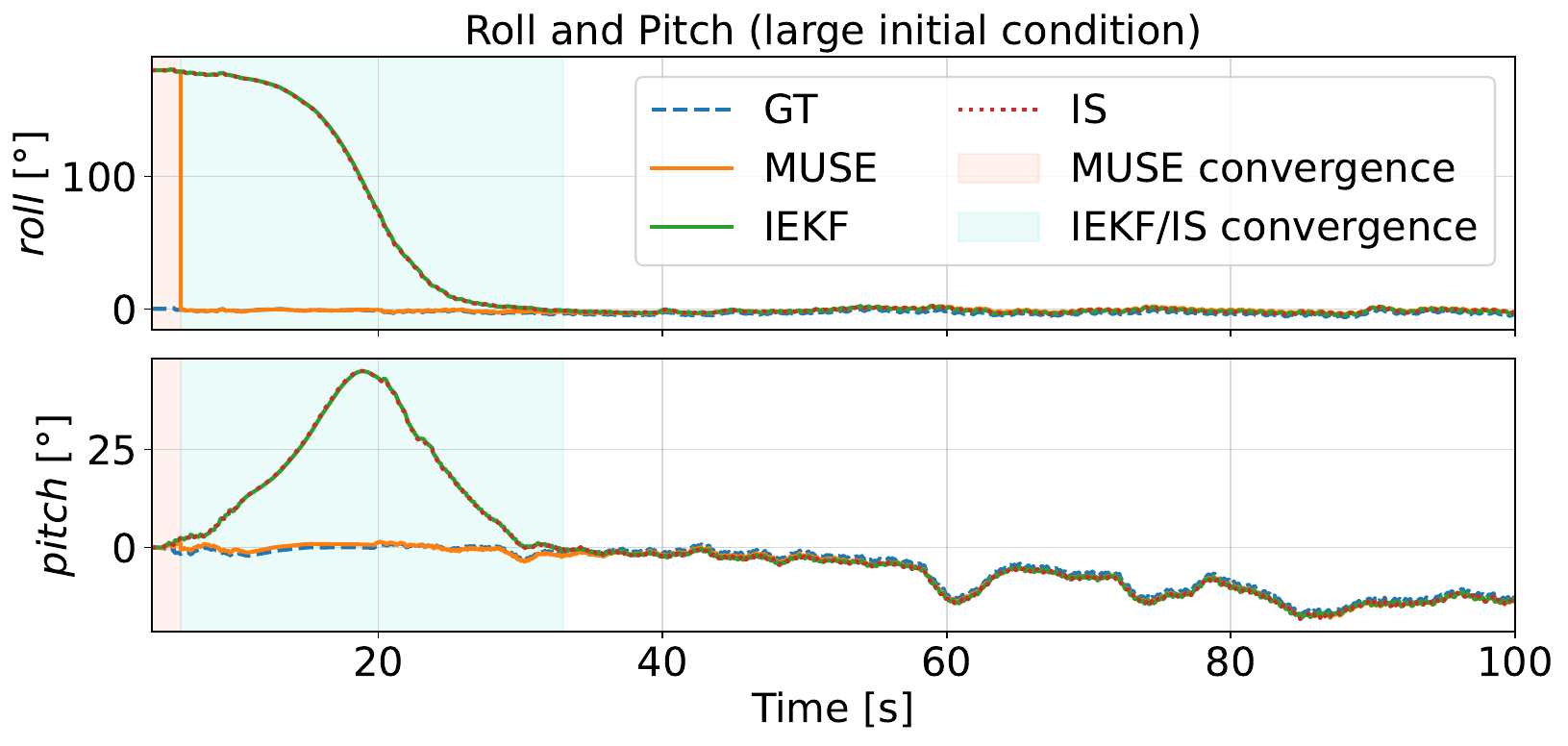}
    \caption{Convergence behavior with large initial orientation errors for MUSE, IEKF, and IS (WS:3). The plot shows the orientation error (in degrees) over time, starting from a large initial misalignment.}
    \label{fig:bad_init}
    \vspace{-0.45cm}
\end{figure}

We also tested the convergence behavior of the three estimators when initialized with large orientation errors. We introduced a significant initial misalignment (180 degrees in roll) and observed how quickly each estimator converged to the correct orientation. The results, shown in \Cref{fig:bad_init}, indicate that all three estimators converge to the correct orientation, but MUSE converges faster than IEKF and IS (WS:3). The reason for this is likely due to the design of MUSE's attitude observer, which is globally stable and can handle large initial errors without diverging. In contrast, IEKF and IS rely on local linearization and filtering assumptions that can be violated when the initial error is large, leading to slower convergence in some cases. 

These results indicate that all three estimators remain robust under large initialization errors, with MUSE exhibiting faster convergence than IEKF and IS. This property is particularly relevant for real-world deployments, where the initial state may be uncertain or perturbed.
In~\Cref{fig:bad_init} we reported only roll and pitch angles, as the yaw angle is not observable from the IMU and kinematics measurements alone, and its convergence behavior in this scenario is not representative of the estimators' performance.

\section{Conclusions }
\label{sec:conclusions}
This work compares three state-of-the-art proprioceptive estimators (MUSE, IEKF, and IS) on a common dataset. ATE and rotational RPE are similar across methods, with IS achieving the best results, while translational RPE shows that at $\Delta=1$~m, IEKF achieves lower error, whereas at $\Delta=1$~frame the gap with MUSE narrows. MUSE and IEKF run at $\approx$0.012--0.020 ms/iteration, while IS exposes an accuracy--latency trade-off via its window size.
These results highlight the trade-off between local consistency, global accuracy, and latency: filters favor high-rate control, while IS with larger windows improves trajectory accuracy at the cost of latency, with small windows providing a compromise.

We hope this preliminary comparison can help practitioners select an estimator matched to their accuracy-latency envelope and motivate broader, standardized benchmarks for quadruped proprioceptive state estimation.
Future work includes evaluation across robots, terrains, and dynamic conditions, as well as closed-loop experiments assessing control performance (e.g., tracking and disturbance rejection) and real-time deployment on embedded hardware to measure latency, load, and power.

\addtolength{\textheight}{0cm}   



\onecolumn
\section{SUPPLEMENTARY MATERIAL}
\label{sec:supp_material}

\begin{figure}[h]
    \centering
    \begin{tikzpicture}[x=1cm,y=1cm,>=stealth]
        \tikzstyle{box}=[draw, rounded corners=2pt, align=center, minimum height=0.8cm, inner sep=2pt]
        \node[box, fill=orange!35, minimum width=2.5cm] (data) at (0,0) {\footnotesize Input Data\\\footnotesize(from the Dataset)\\\footnotesize \texttt{sensor\_data.csv}\\\footnotesize \texttt{groundtruth.csv}};
        \node[box, fill=green!25, minimum width=2.2cm] (prep) at (3.2,0) {\footnotesize Precompute\\\footnotesize Kinematics};
        \node[box, fill=pink!40, minimum width=3.5cm] (est) at (7.1,0) {\footnotesize Estimators\\\footnotesize MUSE \textbar\ IEKF \textbar\ IS};
        \node[box, fill=cyan!20, minimum width=2.8cm] (eval) at (11.4,0) {\footnotesize Evaluation\\\footnotesize TUM + \texttt{evo}\\\footnotesize ATE/RPE/Vel RMSE};
        \draw[->, thick] (data.east) -- (prep.west);
        \draw[->, thick] (prep.east) -- (est.west);
        \draw[->, thick] (est.east) -- (eval.west);
    \end{tikzpicture}
    \caption{Compact supplementary workflow: dataset processing, estimator execution, and metric evaluation.}
    \label{fig:supp_pipeline}
\end{figure}
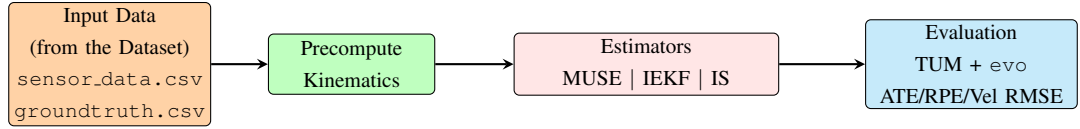

The complete benchmarking pipeline used to produce all results in this paper is available at \url{https://github.com/iit-DLSLab/state_estimation_benchmark}.
The repository provides a self-contained, offline pipeline for quadruped state estimation benchmarking. All three estimators, MUSE, IEKF, and IS, are implemented as standalone C++ executables reading from a shared CSV dataset. Evaluation metrics (ATE, RPE, velocity RMSE) are computed with the \texttt{evo} package~\cite{grupp2017evo} after converting outputs to TUM format~\cite{sturm2012benchmark}. A compact pipeline overview is shown in \Cref{fig:supp_pipeline}, and the full reproducibility details are listed in \Cref{tab:supp}.

\begin{table}[h]
\renewcommand{\arraystretch}{1.3}
\caption{Description of Supplementary Material}
\label{tab:supp}
\centering
\begin{tabular}{p{0.20\textwidth} p{0.74\textwidth}}
\toprule
\textbf{Component} & \textbf{Description} \\
\midrule
\multicolumn{2}{l}{\textit{Repository structure}} \\
\midrule
\texttt{common/}             & Shared data types and CSV I/O utilities (timestamps, IMU, joint states, contacts, ground truth). \\
\texttt{data\_process/}      & Build-and-run tools for dataset inspection and offline precomputation of foot kinematics (positions, Jacobians, velocities) using Pinocchio; Python scripts for plotting, TUM conversion, metric computation, and optional GNSS georeferencing to UTM. \\
\texttt{muse/}               & C++ implementation of the MUSE proprioceptive pipeline~\cite{nistico2025muse} (adapted from \url{https://github.com/iit-DLSLab/muse}), runnable as a single executable (\texttt{main\_muse}) or as three separate modules (\texttt{main\_attitude\_estimation}, \texttt{main\_leg\_odometry}, \texttt{main\_sensor\_fusion}). \\
\texttt{iekf/}               & C++ implementation of the Invariant EKF~\cite{hartley2020contact}, wrapping the InEKF library (\url{https://github.com/RossHartley/invariant-ekf}) with a quadruped-specific measurement interface, run via \texttt{main\_iekf}. \\
\texttt{invariant\_smoother/}& C++ implementation of the fixed-lag Invariant Smoother~\cite{yoon2024invariant} (adapted from \url{https://github.com/DrcdKAIST/invariant_smoother}), with configurable window size, run via \texttt{main\_invariant\_smoother}. \\
\texttt{data/}               & Dataset root: the \href{https://grand-tour.leggedrobotics.com/dataset#mission-14-on}{ANYmal-D CYN-1} sequence from the GrandTour dataset~\cite{frey2026grandtour} (\texttt{sensor\_data.csv}, \texttt{groundtruth.csv}), generated from the publicly available rosbags.\footnotemark[1] \\
\texttt{models/}             & ANYmal-D URDF model used by Pinocchio for forward kinematics.\footnotemark[1] \\
\midrule
\multicolumn{2}{l}{\textit{Benchmarking pipeline} (each step builds independently with CMake)} \\
\midrule
S1 -- Inspect dataset        & Verify CSV parsing, timestamps, and sample counts (\texttt{inspect\_dataset}). \\
S2 -- Precompute kinematics  & Run Pinocchio offline to compute per-foot positions, Jacobians, and velocities for all leg-based estimators (\texttt{precompute\_feet\_kinematics}); output: \texttt{feet\_kinematics.csv}. \\
S3 -- Run estimators         & Execute \texttt{main\_muse}, \texttt{main\_iekf}, and \texttt{main\_invariant\_smoother} (window size configurable); each writes \texttt{fused\_state.csv} to its own subfolder under \texttt{data/}. \\
S4 -- Plot trajectories      & Compare GT vs.\ estimated position, orientation, and velocity traces with \texttt{data\_process/scripts/plots.py}. \\
S5 -- Convert to TUM         & Run \texttt{convert\_to\_tum.py} to produce TUM-format trajectory files; the script also prints ready-to-run \texttt{evo} commands for ATE and RPE evaluation. \\
S6 -- Georeference (opt.)    & Align estimated trajectories to UTM coordinates using GNSS anchor points via \texttt{georef\_from\_gnss.py}. \\
\midrule
\multicolumn{2}{l}{\textit{Dependencies and reproducibility}} \\
\midrule
\texttt{environment.yml}     & \texttt{conda-forge} environment file to install all C++ and Python dependencies consistently: \texttt{conda env create -f environment.yml \&\& conda activate state\_est\_bench}. Key dependencies: CMake\,$\geq$\,3.22, Ninja, Eigen, Pinocchio, \texttt{fmt}, \texttt{yaml-cpp} (C++); NumPy, Pandas, Matplotlib, SciPy, \texttt{pyproj}/PROJ, \texttt{contextily}, \texttt{evo} (Python). \\
\bottomrule
\end{tabular}
\end{table}

\footnotetext[1]{The dataset and model entries are not fixed requirements of the pipeline. They may be replaced by any dataset, model, or sequence that provides the equivalent information needed for estimation and evaluation.}



\twocolumn
\bibliographystyle{IEEEtran}
\bibliography{IEEEexample}

\end{document}